\pdfoutput=1

\documentclass[11pt]{article}

\usepackage{acl}

\usepackage{times}
\usepackage{latexsym}

\usepackage[T1]{fontenc}

\usepackage[utf8]{inputenc}

\usepackage{microtype}

\usepackage{inconsolata}

\usepackage{enumitem}
\usepackage{comment}
\usepackage{graphicx}
\usepackage{multirow}
\usepackage{bm}

\newlist{myitemize}{description}{10}
\setlist[myitemize]{labelindent=4pt, leftmargin=0pt, align=left, noitemsep, topsep=0pt}

\newcommand{\eg}{\textit{e.g.}}
\newcommand{\ie}{\textit{i.e.}}
\newcommand{\kd}{\textsc{kd}}
\newcommand{\mse}{\textsc{mse}}
\newcommand{\ce}{\textsc{ce}}
\newcommand{\nlp}{\textsc{nlp}}
\newcommand{\mrpc}{\textsc{mrpc}}
\newcommand{\mnli}{\textsc{mnli}}
\newcommand{\pawsx}{\textsc{paws-x}}
\newcommand{\xnli}{\textsc{xnli}}
\newcommand{\squad}{\textsc{sq}u\textsc{ad}}
\newcommand{\nq}{\textsc{n}atural \textsc{q}uestions}
\newcommand{\mlqa}{\textsc{mlqa}}
\newcommand{\tydiqa}{\textsc{t}y\textsc{d}i\textsc{qa}}
\newcommand{\ontonotes}{\textsc{o}nto\textsc{n}otes}
\newcommand{\conll}{\textsc{c}o\textsc{nll}}
\newcommand{\klue}{\textsc{klue3}}
\newcommand{\wmtde}{\textsc{wmt16 de-en}}
\newcommand{\wmtro}{\textsc{wmt16 ro-en}}
\newcommand{\roberta}{\textsc{r}o\textsc{bert}a}
\newcommand{\mbart}{m\textsc{bart}}
\newcommand{\hst}{$t_{hs}$}
\newcommand{\llt}{$t_{ll}$}
\newcommand{\boldhst}{$\bm{t_{hs}}$}
\newcommand{\boldllt}{$\bm{t_{ll}}$}
\NewDocumentCommand{\avi}
{ mO{} }{\textcolor{red}{\textsuperscript{\textit{Avi}}\textsf{\textbf{\small[#1]}}}}
\title{An Empirical Investigation into the Effect of \\Parameter Choices in Knowledge Distillation}

\author{Md Arafat Sultan, Aashka Trivedi, Parul Awasthy, Avirup Sil\\ IBM Research AI\\ \texttt{\{arafat.sultan,aashka.trivedi\}@ibm.com}\\ \texttt{\{awasthyp,avi\}@us.ibm.com}}

\begin{document}
\maketitle

\begin{abstract}
We present a large-scale empirical study of how choices of configuration parameters affect performance in knowledge distillation (\kd{}).
An example of such a \kd{} parameter is the measure of distance between the predictions of the teacher and the student, common choices for which include the mean squared error (\mse{}) and the \textsc{kl}-divergence.
Although scattered efforts have been made to understand the differences between such options, the \kd{} literature still lacks a systematic study on their general effect on student performance.
We take an empirical approach to this question in this paper, seeking to find out the extent to which such choices influence student performance across $13$ datasets from $4$ \nlp{} tasks and $3$ student sizes.
We quantify the cost of making sub-optimal choices and identify a single configuration that performs well across the board.
\end{abstract}

\section{Introduction}
\label{section:introducion}
Knowledge distillation (\kd{}) from a larger \textit{teacher} model into a smaller \textit{student} model \cite{hinton2014distilling} is a powerful approach in machine learning, often yielding better models than ordinary 1-hot-label training.
Yet setting up \kd{} and optimizing its performance for a new application can be challenging due to its highly parameterized nature.
For example, one can measure the distance between the predictions of the teacher and the student as the mean squared error (\mse{}) between their output logits, or alternatively with cross-entropy or \textsc{kl}-divergence between their probability estimates \cite{kim2021comparing}.
The teacher, as another example, can be selected based on its performance according to some evaluation metric or the quality of its probability estimates \cite{menon2021statistical}.

Crucially, our understanding of the extent to which the various available options in \kd{} represent fundamentally different processes, and not minor variations of the same underlying mechanism with similar performance profiles, is still very limited.
As a result, users of \kd{} often resort to personal experience (and bias) to choose from such alternatives or even do so randomly \citep{mukherjee2020xtremedistil, sanh2020distilbert, jiao2020tinybert, wang2021selective, zhou-etal-2021-multi}, not knowing how their choices might affect downstream results.
In this paper, we take a first step towards understanding the effects of different parameter choices on the performance of \kd{} by conducting a large-scale empirical investigation on $13$ carefully chosen datasets (\S{\ref{section:methodology}})
from $4$ \nlp{} tasks
with models of different sizes and capacities.

Besides the aforementioned distance measure and teacher selection criterion, we include two more key \kd{} parameters in our study (\S{\ref{section:preliminaries}}): whether or not to (\textit{a})~use human labels for supervision, and (\textit{b})~temperature scale the student's output.
We perform an approximate greedy search over the large space of combinations of these choices (\S{\ref{section:methodology}}), and present empirical results answering the following research questions (\S{\ref{section:experiments-and-results}}): \textbf{(\textit{i})~How much do parameter choices matter in \kd{}?} We quantify and compare performances of strong, weak and random \kd{} configurations; 
\textbf{(\textit{ii})~What are the effects of individual parameters?} We examine if certain parameters play a bigger role in determining student performance; and \textbf{(\textit{iii})~Does a single configuration exist that can perform well across the board?} We propose a simple method to find such a combination of parameter choices using validation set results and analyze its test set performance.

Our results show a relative performance gain of up to $9.4\%$ with a strong student model over a weak one ($4.3\%$ in the 90th percentile), and that of up to $3\%$ over a randomly chosen configuration.
Since our search over \kd{} configurations is not exhaustive (\S{\ref{section:methodology}}), these are in fact lower bounds of the corresponding quantities, indicating that a sub-optimal choice may lead to undesired, even critical, performance loss for many applications.
To that end, the one configuration identified by our proposed method not only reduces the performance gap across the board, but also matches or outperforms the original best configurations -- which are test-specific and variable -- in $40\%$ of all test cases.

\section{Preliminaries}
\label{section:preliminaries}
Given input $x$, let $q(x)$ be the 1-hot target distribution to learn in a task.
Let $l^s(x)$ and $p^s(x)$ be the output logits and the corresponding softmax distribution predicted by the student $s$ being trained, and $l^t(x)$ and $p^t(x)$ the same for the pre-trained teacher $t$.
The \kd{} objective for $x$ takes the following general form (omitting arguments for simplicity):
\begin{equation}
\footnotesize
\label{equation:kd-cost-function}
    \mathcal{L}_{KD} = (1 - \alpha) \mathcal{L}_1(q,p^s) + {\alpha} \mathcal{L}_2(o^t_{\tau_t},o^s_{\tau_s})
\end{equation}
where $\mathcal{L}_1$ is the distance of $p^s$ from $q$ and $\mathcal{L}_2$ is the distance between the output $o^t$ of the teacher and $o^s$ of the student, which can be either the logits $l^t$ and $l^s$ or the probabilities $p^t$ and $p^s$ depending on the choice of $\mathcal{L}_2$, with $\alpha \in (0, 1]$ modulating the weights of $\mathcal{L}_1$ and $\mathcal{L}_2$ in the overall loss.
$\tau_t$ and $\tau_s$ are temperature hyperparameters that can be used to divide the respective sets of logits to smoothen the corresponding distributions.

We investigate the following decision choices involving four key \kd{} parameters:
\begin{myitemize}
\item[$\bullet$]
\textbf{\textit{Use of human labels.}} Setting $\alpha=1$ in Eq.~\ref{equation:kd-cost-function} uses only the teacher's predictions as targets for distillation \cite{jiao2020tinybert, beyer2022good}; alternatively, one can tune $\alpha\in(0,1]$ to also learn from 1-hot expert labels \cite{sun2019patient, dasgupta-etal-2023-cost}. There is an important trade-off associated with this choice: tuning $\alpha$ can 
increase the number of training runs by a factor of $n_\alpha$, where $n_\alpha$ is the number of $\alpha$ values to search over.
\item[$\bullet$] \textbf{\textit{Teacher-student distance measure.}} We compare performances of two $\mathcal{L}_2$ options in Eq.~\ref{equation:kd-cost-function}: the mean-squared error (\textbf{\mse{}}) between $l^s$ and $l^t$ \cite{kim2021comparing, rao2023dynamic} and the cross-entropy (\textbf{\ce{}}) between $p^s$ and $p^t$ \cite{hinton2014distilling, Yang_Pan_Gao_Jiang_Liu_Chen_2022}. 
A related study by \citet{kim2021comparing} discusses empirical observations in image classification, but our more extensive investigation includes a larger set of \nlp{} tasks.
\item[$\bullet$] \textbf{\textit{Teacher selection.}} \citet{menon2021statistical} find that models that generate high-quality probability estimates -- as indicated by their low log-loss on a held-out set -- constitute better \kd{} teachers than models with a high top-1 accuracy. While they compare teachers $t$ of different architectures and sizes, our goal here is to identify the best checkpoint for $t$ stored during training. For each teacher-student-dataset triple, we compare students distilled from two teachers: (\textit{a})~the highest-score teacher (\boldhst{}): the teacher checkpoint with the highest validation set score for the chosen evaluation metric, \eg{}, F1-score, and (\textit{b})~the lowest loss teacher (\boldllt{}): the checkpoint with the lowest validation set loss.
\item[$\bullet$] \textbf{\textit{Temperature scaling of student logits.}} While \citet{hinton2014distilling} originally applied an equal amount of temperature scaling to teacher and student outputs ($\tau_s=\tau_t$), some later work scaled only the teacher's ($\tau_s=1$) \cite{zhang2020self,sultan2022overfit}. The latter has been found to generate better calibrated predictions \cite{zhang2020self} and is arguably easier to interpret; here we investigate if it also leads to significant performance differences across a range of tasks.
\end{myitemize}

\section{Methodology}
\label{section:methodology}

\textbf{\textit{Setting up a diverse set of test conditions.}}
A study such as this must involve a variety of test cases with a multitude of functions to learn as well as models of different sizes and capacities.
To that end, we 
evaluate \kd{} parameter choices on four \nlp{} tasks: text classification, reading comprehension, named entity recognition and machine translation.
The first three are extractive or labeling tasks with different label sets and the fourth a generative task.
To further enforce diversity, within each task, we select datasets that vary in properties such as
the size of their training sets and the mix of mono and multilingual data in their splits.
Table~\ref{table:tasks-and-datasets} lists the datasets; see Appendix~\ref{section:appendix-datasets} for details.

\begin{table}[t]
    \scriptsize
    \centering
    \begin{tabular}{l|c}
         \multicolumn{1}{c|}{\textbf{Task}} & \textbf{Datasets} \\
         \hline
         Text Classification & \mrpc{}, \mnli{}, \pawsx{}, \xnli{} \\
         \hline
         Named Entity Recognition & \ontonotes{}, \conll{}, \klue{} \\
         \hline
         \multirow{2}{*}{Reading Comprehension} & \squad{}, \nq{}, \\
         & \mlqa{}, \tydiqa{} \\
         \hline
         \multirow{2}{*}{Machine Translation} & \wmtde{}, \\
         & \wmtro{} \\
    \end{tabular}
    \caption{The $4$ tasks and $13$ datasets used in our experiments. See Appendix~\ref{section:appendix-datasets} for details.}
    \vspace{-2mm}
    \label{table:tasks-and-datasets}
\end{table}

For each non-generative task, we fine-tune a \roberta{}-large \cite{liu2019roberta} teacher for each dataset and distill it into a \roberta{}-base as well as two $6$- and $4$-layer truncated \roberta{} architectures.
The two latter students are initialized using a skip-layer strategy as in \citet{sun2019patient}.
For machine translation, we similarly fine-tune an \mbart{}-large-cc25 \cite{liu2020multilingual} teacher ($12$ encoder and $12$ decoer layers) and distill it into $12$-, $6$- and $4$-layer \mbart{}-cc25 skipped-layer students, each with an equal number of encoder and decoder layers.

\begin{table}[t]
    \footnotesize
    \centering
    \begin{tabular}{c|c}
        \textbf{Parameter} & \textbf{Options} (\S{\ref{section:preliminaries}}) \\
        \hline
        $t$ & \hst{}, \llt{} \\
        $\mathcal{L}_2$ & \ce{}, \mse{} \\
        $\tau_s$ & $1, \tau_t$ \\
        $\alpha$ & \{1\}, $\{.1, .2, ..., .9, 1\} $ \\
    \end{tabular}
    \caption{\kd{} parameter grid in our experiments.}
    \label{table:kd-parameters}
    \vspace{-2mm}
\end{table}

\noindent \textbf{\textit{Searching over a large space of parameter choices.}}
Table~\ref{table:kd-parameters} shows a typical parameter search space for a single \kd{} experiment.
Two of the four parameters in the table ($t$ and $\mathcal{L}_2$) are categorical and one ($\alpha$) is bounded; for these, a standard grid search over a fixed set of values is sufficient.
$\tau_t$, however, is unbounded, and thus a two-stage search, first over a sparse set of values, \eg{}, a geometric progression, and then linearly within a local proximity of the optimal value from stage $1$ arguably makes more sense for it (and for $\tau_s$ when $\tau_s=\tau_t$).

Assuming a grid size of $7$ for the first broad search for $\tau_t$ and of $4$ for the second local search, the number of models to be distilled and evaluated per dataset-student pair implied by Table~\ref{table:kd-parameters} alone is roughly $440$; this number must be multiplied by the size of the space of hyperparameters associated with general model training, \eg{}, learning rate, to compute the actual number of students to be trained.
This shows why an exhaustive grid search is practically infeasible in \kd{}, not only for practitioners in general but also for us in this study, as we intend to test the effects of such choices across different tasks, datasets and model sizes.

We therefore adopt the following greedy search strategy to navigate important points of the grid:
First, we set $\alpha$ to $1$ -- training the student with only the teacher's soft predictions -- and optimize the remaining parameters $\mathcal{L}_2$, $t$ and $\tau_s$; we then evaluate this optimal 
$\alpha=1$ combination of the other three parameters for additional values of $\alpha$ (Table~\ref{table:kd-parameters}) to determine if a mix of hard and soft labels can train better students than only the latter.
To find the best combination of $\mathcal{L}_2$, $t$ and $\tau_s$ for $\alpha=1$, we start with the configuration of \citet{hinton2014distilling} ($\mathcal{L}_2=$ \ce{}, $t=$ \hst{}, $\tau_s=\tau_t$) and tune the learning rate to find the best model 
using the validation set, and evaluate it on the test set.
Then at each step of an iterative process, we change the value of one \kd{} parameter and evaluate it similarly.
The value of the parameter is set to the choice that yields the higher score and the process continues.
This method traverses four out of eight possible configurations for $\alpha=1$ with an aim to evaluate the more promising candidates.

\section{Experimental Results}
\label{section:experiments-and-results}
We compare different \kd{} configurations using the following uniform method in all our evaluations: (\textit{i})~Compute the relative gain in accuracy with a target configuration over a baseline for all $39$ teacher-student-dataset triples, \eg{}, (\squad{}, \roberta{}-large, \roberta{}-4layer), (\textit{ii})~Split the $39$ gains up into deciles, and (\textit{iii})~Put different such comparisons side by side to assess their magnitude in context.
The Transformers library of \citet{wolf2020transformers} is used for all implementations.

\begin{figure}[t]
\includegraphics[scale=0.5]{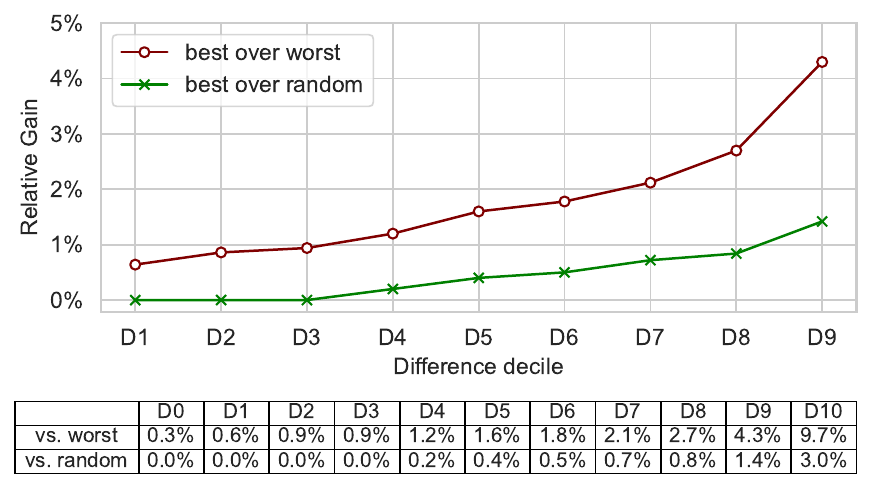}
\centering
\caption{
Relative performance gain with the best \kd{} configuration in our evaluated sample over two baselines.
The empirical upper bound of the cost of making a \textit{bad} parameter choice is $1.6\%$ with a $50\%$ probability and $4.3\%$ with a $90\%$ probability.
}
\label{figure:opitmal-v-baselines}
\vspace{-2mm}
\end{figure}

Figure~\ref{figure:opitmal-v-baselines} illustrates how the best \kd{} configuration for each individual dataset-teacher-student triple, \ie{} \textit{test case}, improves performance over two baselines: the worst-performing configuration for the triple and a configuration chosen randomly from among all candidates.
At the median ($D5$), we observe a gain of $0.4$--$1.6\%$ over the baselines, which goes up to $1.4$--$4.3$\% in the $9$th decile ($D9$).
In other words, the empirical upper bound of the cost of making a \textit{bad} parameter choice is $1.6\%$ with a $50\%$ probability and $4.3\%$ with a $90\%$ probability.
The interpretation of how big these differences are can vary from one use case to another: the $3\%$ difference at the peak between a tuned and a random configuration in Figure~\ref{figure:opitmal-v-baselines} likely does not warrant any expensive tuning in, for example, an academic research project where \kd{} is being used simply as a tool among many and is not the central research topic.
In contrast, an ill-fated worse-case scenario causing a near $10\%$ drop, as we observe for $D10$ of the worst-performing configuration, can be critical in many real-life applications.
Given that we only evaluate a subset of all configurations (\S{\ref{section:methodology}}) -- and thus the differences we observe in Figure~\ref{figure:opitmal-v-baselines} are in fact lower bounds -- \kd{} configuration choices can have far-reaching consequences in practice.

\begin{figure}[t]
\includegraphics[scale=0.48]{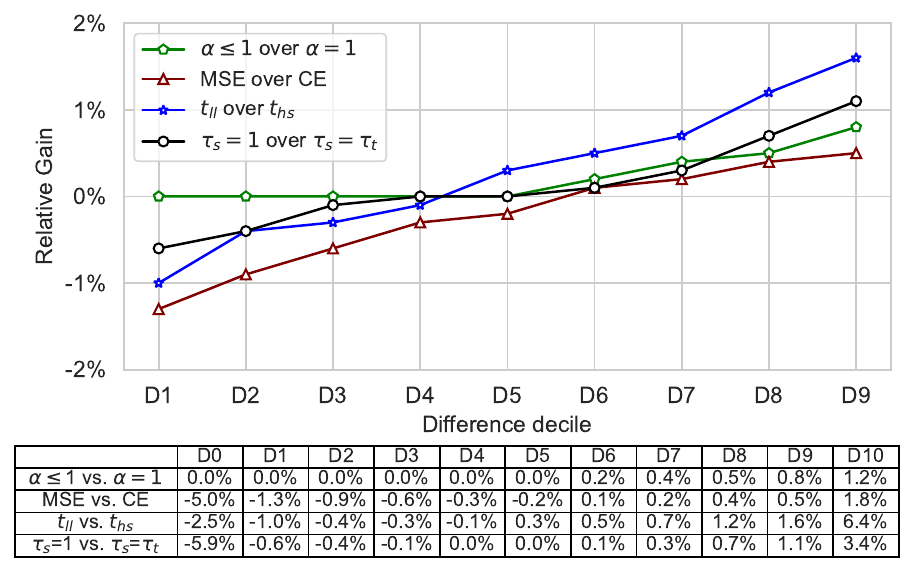}
\centering
\caption{
Performance differences due to individual \kd{} parameter choices.
Choices of all parameters except $\alpha$ can have a non-negligible impact on performance.
} 
\label{figure:importance-of-individual-params}
\vspace{-2mm}
\end{figure}

To gain further insight, next we look at the impact of individual parameter choices.
As described in \S{\ref{section:methodology}}, at each step of our greedy grid search for $\alpha=1$, we change the value of one of the other parameters $t$, $\mathcal{L}_2$ and $\tau_s$ and evaluate the resulting configuration; we take the performances of the two configurations before and after this change as an estimate of the performance of the two corresponding parameter values.
In Figure~\ref{figure:importance-of-individual-params}, the use of hard labels ($\alpha\leq1$) has the least overall impact over its baseline ($\alpha=1$), which is an important finding as tuning $\alpha$ can be expensive. 
For the remaining parameters, the difference in student performance has a much wider distribution over the two choices, with \ce{} as a distance measure perhaps showing the clearest advantage over \mse{}. 
Individual choices make less of a difference than their combination, as expected, when compared to Figure~\ref{figure:opitmal-v-baselines}.

\begin{table}[h]
    \scriptsize
    \centering
    \begin{tabular}{c|cc|c}
        \multicolumn{1}{c|}{\textbf{Parameter}} & \textbf{Default} & \textbf{Alternative} & $\bm{R}$ \\
        \hline
        $t$ & \hst{} & \boldllt{} & $+11$ \\
        $\mathcal{L}_2$ & \textbf{\ce{}} & \mse{} & $-5$ \\
        $\tau_s$ & $\tau_t$ & $\bm{1}$ & $+1$
    \end{tabular}
    \caption{Rewards $R$ for overriding default parameter choices with their alternatives.
    The best configuration implied by this table is thus: $t=$ \llt{}, $\mathcal{L}_2=$ \ce{}, $\tau_s=1$.
    }
    \vspace{-2mm}
    \label{table:individual-parameter-choices-on-dev}
\end{table}

Our final research question concerns finding a single \kd{} configuration that is stable, \ie{}, performs well across different applications.
Given the small observed improvement from tuning $\alpha$, we explore this question only for $\alpha=1$ and look for a stable combination of $t$, $\mathcal{L}_2$ and $\tau_s$ values.
Assuming mutual independence of these parameters, we separately optimize for each on validation data, and evaluate the combination of optimal values as our target configuration on all test sets.
Table~\ref{table:individual-parameter-choices-on-dev} shows the validation set reward $R$ for overriding the \citet{hinton2014distilling} choice (the \textit{default}) for each parameter with the alternative choice, which we compute as the difference between the number of times the alternative trains a better and a worse student than the default across all $39$ of our test cases.
A positive $R$ 
implies that the alternative choice should be favored 
while
$R<0$
indicates that the default is the better option; the magnitude of $R$ represents the strength of the finding.
Our results confirm the superiority of the low-loss teacher of \citet{menon2021statistical} in a larger context, and shows $\tau_s=1$ to be on par with the traditional $\tau_s=\tau_t$, which is desirable given its better known calibration properties \cite{zhang2020self}.

\begin{figure}[t]
\includegraphics[scale=0.5]{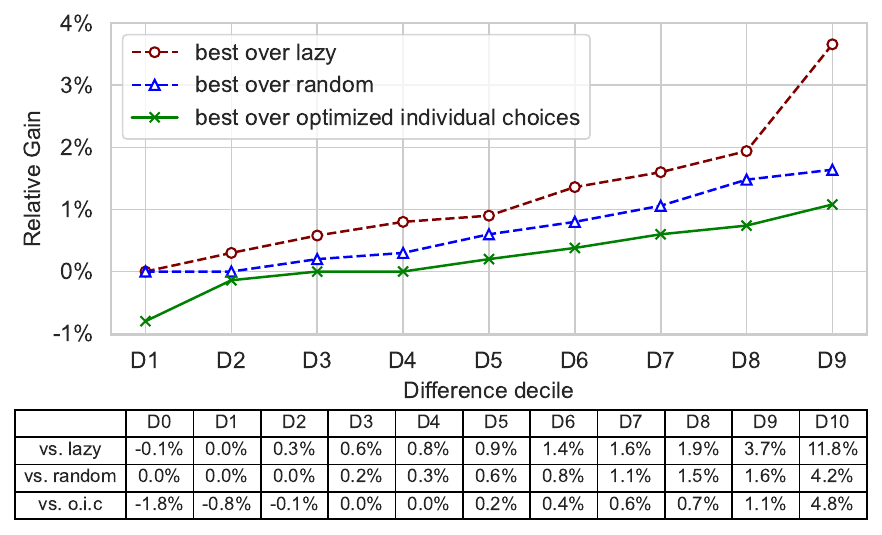}
\centering
\caption{Our proposed configuration (green) optimized over validation data performs at least as well as the original test-specific best configurations in $40\%$ of all test cases, outperforming two baselines across the board.
}
\label{figure:optimal-v-gbic}
\vspace{-2mm}
\end{figure}

In Figure~\ref{figure:optimal-v-gbic} we contrast test set performances of the best configuration of Table~\ref{table:individual-parameter-choices-on-dev} with those of a random and a \textit{lazy} configuration; the latter simulates a common user persona who chooses a random \kd{} configuration along with $\tau_t=1$, as in \citet{ren2023tailoring}.
Our best single configuration shows the least lag from the best-performing configuration specific to each individual test case; in fact, it is at least as good as the latter in $40\%$ of all test cases and better than both baselines across the board.
This is yet another significant discovery of our study, pointing to a relatively stable \kd{} configuration if no tuning were to be performed.

\section{Conclusion}
\label{section:conclusion}
We demonstrate substantial differences in student accuracy due to varying choices of configuration parameters in knowledge distillation (\kd{}), and identify a single configuration that stabilizes performance across a range of different target scenarios.
While we only take a small, yet crucial, step towards understanding the effects of parameter choices in \kd{}, through this work, we hope to inspire future explorations of this important topic.

\section*{Limitations}
\label{section:limitations}
First, due to a prohibitively large search space, we were able to perform only an approximate grid search over \kd{} parameters to find our best and worst empirical configurations (\S{\ref{section:methodology}}).
Second, even with a large-scale investigation involving a variety of test conditions, it is generally hard to claim with certainty that the findings of such a study will generalize to even larger studies with more tasks.
However, scientific knowledge often advances one small step at a time, which is how we believe the contributions of this work should be viewed.

\bibliography{custom}

\appendix

\section{Evaluation Metrics}
\label{section:appendix-metrics}
We use standard evaluation metrics for all tasks: \textsc{bleu} for machine translation and F1-score for the rest.

\section{Details of Datasets}
\label{section:appendix-datasets}
Our datasets represent a diverse set of tasks, varying in properties such as size, language and label space.
In this section, we describe key attributes of all $13$ of our datasets.

\subsection{Text Classification}
\begin{itemize}
\item \textbf{\mrpc{}}~\cite{dolan2005mrpc}:
\begin{itemize}
    \item $\#$ of training examples: $3,668$
    \item $\#$ of validation examples: $408$
    \item $\#$ of test examples: $1,725$
    \item Training set language: English
    \item Validation set language : English
    \item Test set language: English
    \item $\#$ of labels: 2
\end{itemize}
\item \textbf{\mnli{}}~\cite{williams2018mnli}:
\begin{itemize}
    \item $\#$ of training examples: $382,887$
    \item $\#$ of validation examples: $9,815$
    \item $\#$ of test examples: $9,815$
    \item Training set language: English
    \item Validation set language : English
    \item Test set language: English
    \item $\#$ of labels: 3
\end{itemize}
\item \textbf{\pawsx{}}~\cite{yang2019pawsx}:
\begin{itemize}
    \item $\#$ of training examples: $345,807$
    \item $\#$ of validation examples: $14,000$
    \item $\#$ of test examples: $14,000$
    \item Training set language: Multilingual
    \item Validation set language : Multilingual
    \item Test set language: Multilingual
    \item $\#$ of labels: 2
\end{itemize}
\item \textbf{\xnli{}}~\cite{conneau2018xnli}:
\begin{itemize}
    \item $\#$ of training examples: $392,702$
    \item $\#$ of validation examples: $37,350$
    \item $\#$ of test examples: $75,150$
    \item Training set language: English
    \item Validation set language: Multilingual
    \item Test set language: Multilingual
    \item $\#$ of labels: 3
\end{itemize}

\end{itemize}

\subsection{Reading Comprehension}
\begin{itemize}
\item \textbf{\squad{}}~\cite{rajpurkar2016squad}:
\begin{itemize}
    \item $\#$ of training examples: $76,081$
    \item $\#$ of validation examples: $10,507$
    \item $\#$ of test examples: $10,507$
    \item Training set language: English
    \item Validation set language : English
    \item Test set language: English
    \item $\#$ of labels: $384$ (context window size)
\end{itemize}
\item \textbf{\nq{}}~\cite{kwiatkowski-etal-2019-natural}:
\begin{itemize}
    \item $\#$ of training examples: $91,235$
    \item $\#$ of validation examples: $12,836$
    \item $\#$ of test examples: $12,836$
    \item Training set language: English
    \item Validation set language : English
    \item Test set language: English
    \item $\#$ of labels: $384$ (context window size)
\end{itemize}
\item \textbf{\mlqa{}}~\cite{lewis-etal-2020-mlqa}:
\begin{itemize}
    \item $\#$ of training examples: $76,081$
    \item $\#$ of validation examples: $7,715$
    \item $\#$ of test examples: $75,712$
    \item Training set language: English
    \item Validation set language : Multilingual
    \item Test set language: Multilingual
    \item $\#$ of labels: $384$ (context window size)
\end{itemize}
\item \textbf{\tydiqa{}}~\cite{clark-etal-2020-tydi} (Gold Passages):
\begin{itemize}
    \item $\#$ of training examples: $44,896$
    \item $\#$ of validation examples: $4,985$
    \item $\#$ of test examples: $5,077$
    \item Training set language: Multilingual
    \item Validation set language : Multilingual
    \item Test set language: Multilingual
    \item $\#$ of labels: $384$ (context window size)
\end{itemize}
\end{itemize}

\subsection{Named Entity Recognition}
\begin{itemize}
\item \textbf{\conll{}}~\cite{tjong-kim-sang-2002-introduction}:
\begin{itemize}
    \item $\#$ of training examples: $39,115$
    \item $\#$ of validation examples: $8,275$
    \item $\#$ of test examples: $10,395$
    \item Training set language: Multilingual
    \item Validation set language: Multilingual
    \item Test set language: Multilingual
    \item $\#$ of labels: $9$ 
\end{itemize}
\end{itemize}

\begin{itemize}
\item \textbf{\ontonotes{}}~\cite{pradhan-etal-2012-conll}:
\begin{itemize}
    \item $\#$ of training examples: $82,728$
    \item $\#$ of validation examples: $10,510$
    \item $\#$ of test examples: $10,396$
    \item Training set language: English
    \item Validation set language : English
    \item Test set language: English
    \item $\#$ of labels: $37$ 
\end{itemize}
\end{itemize}

\begin{itemize}
\item \textbf{\klue{}}~\cite{florian-etal-2010-improving}:
\begin{itemize}
    \item $\#$ of training examples: $12,467$
    \item $\#$ of validation examples: $8,785$
    \item $\#$ of test examples: $10,816$
    \item Training set language: English
    \item Validation set language: Multilingual
    \item Test set language: Multilingual
    \item $\#$ of labels: $103$ 
\end{itemize}
\end{itemize}

\subsection{Machine Translation}
\begin{itemize}
\item \textbf{\wmtde{}}~\cite{bojar-etal-2016-findings} (maximum sequence length of 128 tokens):
\begin{itemize}
    \item $\#$ of training examples: $500,000$ (sampled from $4,490,021$)
    \item $\#$ of validation examples: $2,168$
    \item $\#$ of test examples: $2,998$
    \item Training set language: German (source), English (target)
    \item Validation set language: 
    German (source), English (target)
    \item Test set language: German (source), English (target)
    \item $\#$ of labels: $50,265$ (vocabulary size) 
\end{itemize}
\end{itemize}
\begin{itemize}
\item \textbf{\wmtro{}}~\cite{bojar-etal-2016-findings} (maximum sequence length of 128 tokens):
\begin{itemize}
    \item $\#$ of training examples: $100,000$ (sampled from $596,163$)
    \item $\#$ of validation examples: $1,998$
    \item $\#$ of test examples: $1,996$
    \item Training set language: Romanian (source), English (target)
    \item Validation set language: 
    Romanian (source), English (target)
    \item Test set language: Romanian (source), English (target)
    \item $\#$ of labels: $50,265$ (vocabulary size) 
\end{itemize}
\end{itemize}

\end{document}